\documentclass[letterpaper]{article} % DO NOT CHANGE THIS
\usepackage{arxiv}  % DO NOT CHANGE THIS
\usepackage{times}  % DO NOT CHANGE THIS
\usepackage{helvet}  % DO NOT CHANGE THIS
\usepackage{courier}  % DO NOT CHANGE THIS
\usepackage[hyphens]{url}  % DO NOT CHANGE THIS
\usepackage{graphicx} % DO NOT CHANGE THIS
\usepackage{natbib}  % DO NOT CHANGE THIS AND DO NOT ADD ANY OPTIONS TO IT
\usepackage{caption} % DO NOT CHANGE THIS AND DO NOT ADD ANY OPTIONS TO IT
\usepackage{algorithm}
\usepackage{amsmath}   % 支持 equation、align 等数学环境
\usepackage{amssymb}   % 提供 \odot 等数学符号
\usepackage{multirow}
\usepackage[table,xcdraw]{xcolor}
\usepackage{graphicx}     % 支持插入图片
\usepackage{caption}
\usepackage{amssymb} 
\usepackage{mathtools}
\usepackage{amsthm}
\usepackage{adjustbox}
\usepackage{colortbl,xcolor}
\usepackage{algorithmic}

\theoremstyle{plain}

\theoremstyle{definition}

\theoremstyle{remark}

\usepackage[textsize=tiny]{todonotes}
\usepackage{newfloat}
\usepackage{listings}
\DeclareCaptionStyle{ruled}{labelfont=normalfont,labelsep=colon,strut=off} % DO NOT CHANGE THIS
\floatstyle{ruled}
\newfloat{listing}{tb}{lst}{}
\floatname{listing}{Listing}
\title{DiffStyle3D: Consistent 3D Gaussian Stylization via Attention Optimization}
\author{
    Yitong Yang\textsuperscript{\rm 1},
    Xuexin Liu\textsuperscript{\rm 1},
    Yinglin Wang\textsuperscript{\rm 1}\thanks{Corresponding author.},
    Jing Wang\textsuperscript{\rm 1},
    Hao Dou\textsuperscript{\rm 1}, \\
    Changshuo Wang\textsuperscript{\rm 2},
    Shuting He\textsuperscript{\rm 1}\footnotemark[1]
}
\affiliations{
    \textsuperscript{\rm 1}School of Computing and Artificial Intelligence, Shanghai University of Finance and Economics\\
    \textsuperscript{\rm 2}Department of Computer Science, University College London, University of London\\
    yangyitong@stu.sufe.edu.cn, \{wang.yinglin, shuting.he\}@sufe.edu.cn
}

\usepackage{bibentry}

\let\oldtwocolumn\twocolumn
\renewcommand\twocolumn[1][]{%
    \oldtwocolumn[{#1}{
    \begin{center}
        \centering
        \includegraphics[width=1.0\textwidth]{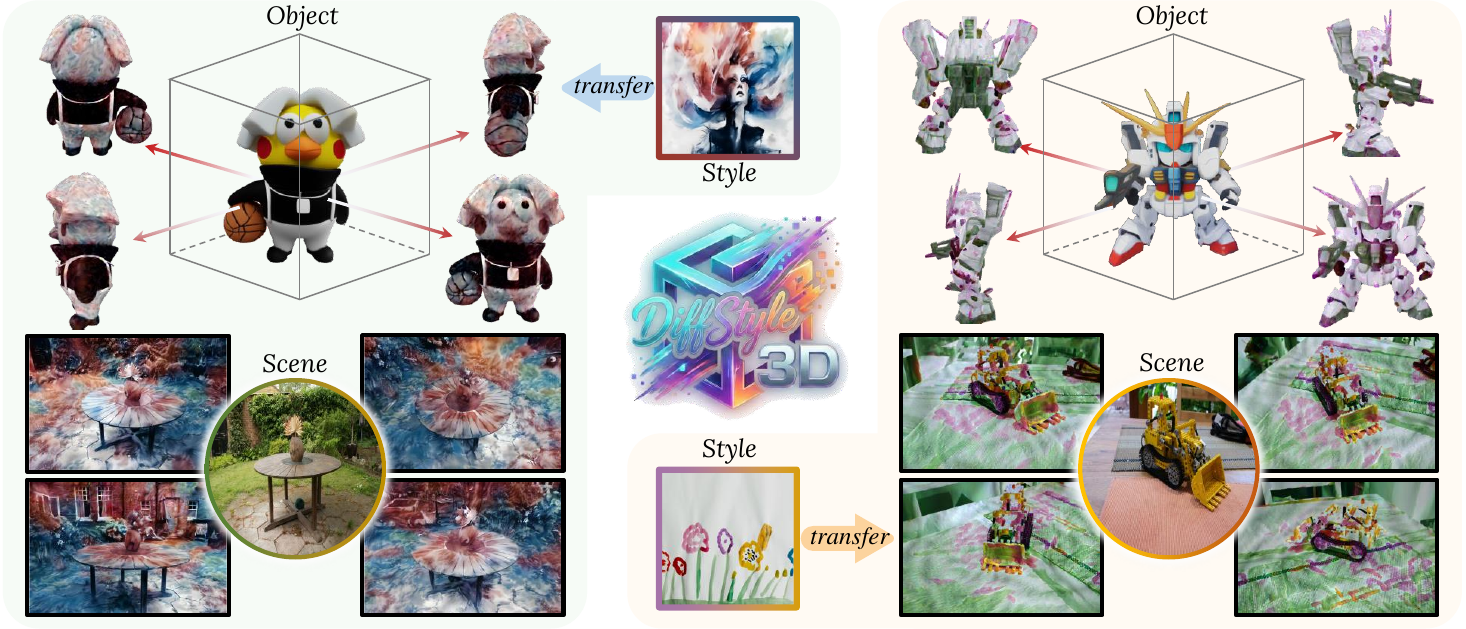}
        \captionof{figure} {Our method enables high-quality 3D stylization across diverse styles for both scenes and objects.}
        \label{fig:teaser}
    \end{center}
    }]
}

\begin{document}
\maketitle

\begin{abstract}
3D style transfer enables the creation of visually expressive 3D content, enriching the visual appearance of 3D scenes and objects. However, existing VGG- and CLIP-based methods struggle to model multi-view consistency within the model itself, while diffusion-based approaches can capture such consistency but rely on denoising directions, leading to unstable training. To address these limitations, we propose DiffStyle3D, a novel diffusion-based paradigm for 3DGS style transfer that directly optimizes in the latent space. Specifically, we introduce an Attention-Aware Loss that performs style transfer by aligning style features in the self-attention space, while preserving original content through content feature alignment. Inspired by the geometric invariance of 3D stylization, we propose a Geometry-Guided Multi-View Consistency method that integrates geometric information into self-attention to enable cross-view correspondence modeling. Based on geometric information, we additionally construct a geometry-aware mask to prevent redundant optimization in overlapping regions across views, which further improves multi-view consistency. Extensive experiments show that DiffStyle3D outperforms state-of-the-art methods, achieving higher stylization quality and visual realism.
\end{abstract}
\section{Introduction}
\label{sec:intro}
% 2D image style transfer~\cite{kwon2022clipstyler,zhang2023inversion,yang2025splitflux} has achieved remarkable progress, significantly enriching visual expression and artistic representation. However, 
With the rapid development of applications such as virtual reality, gaming, and film production, the demand for digital content is shifting from 2D images to 3D representations, making large-scale, high-quality 3D assets increasingly essential~\cite{he2025survey}. Against this backdrop, 3D stylization has emerged as a promising research direction, aiming to transform static 3D geometric representations into expressive digital assets with distinctive aesthetic characteristics, thereby facilitating the low-cost, efficient, and scalable creation of high-quality 3D artistic content. Previous 3D style transfer methods~\cite{chen2024upst,liu2023stylerf,fujiwara2024style,zhang2022arf} predominantly relied on NeRF-based~\cite{mildenhall2021nerf} representations, which suffer from substantial computational overhead and long training times, limiting their efficiency and scalability. Recently, 3D Gaussian Splatting (3DGS)~\cite{kerbl20233d} has emerged as a promising alternative, offering significantly improved rendering efficiency and high visual quality. As a result, 3DGS has quickly become a focal point in 3D style transfer research. 

Currently, 3DGS-based style transfer research~\cite{zhuang2025styleme3d,ArtNVG,lin2025multi,zhang2024stylizedgs} can be broadly divided into three categories. First, VGG-based methods~\cite{StyleGaussian, galerne2025sgsst, lin2025multi, saroha2024gaussian}, inspired by 2D feature statistic matching~\cite{gatys2016image,jing2019neural}, enforce style consistency by minimizing Gram matrix discrepancies. While these methods offer stable training, inherent model limitations hinder multi-view consistency, often resulting in inconsistent stylization across varying viewpoints. Second, CLIP-based methods~\cite{howil2025clipgaussian, kovacs2024g} align feature directions within the CLIP embedding space to introduce semantic style constraints; however, they similarly fail to explicitly model cross-view correspondences, leading to stylistic drift or flickering. Recently, diffusion-based methods~\cite{yang2025fantasystyle} leverage the intrinsic properties of diffusion models to establish multi-view consistency. Nevertheless, these approaches frequently suffer from unstable training and potential artifacts, such as over-smoothing, due to their reliance on optimization along predicted denoising directions.

To address these challenges, we propose DiffStyle3D, a novel diffusion-based paradigm for 3DGS stylization. Unlike previous approaches that rely on denoising directions from diffusion models for optimization, we introduce an Attention-Aware Loss that achieves stable 3D style transfer through direct latent-space optimization while effectively preserving the original content. It consists of two terms. (1) Style loss: within self-attention, we inject the keys (K) and values (V) from the style image into the original queries (Q), using the resulting attention output as a stylization signal to guide the integration of style information into the 3D representation. (2) Content loss: to preserve content fidelity, we align the attention outputs of the content image with those of the rendered image.

Motivated by the geometric invariance of 3D stylization, where only color-related parameters are optimized while geometry remains fixed, we propose Geometry-Guided Multi-View Consistency. By leveraging camera parameters and depth maps, we explicitly determine geometric relationships and incorporate this information into the self-attention mechanism to form Geometry-Guided Attention, thereby modeling cross-view correspondences and mitigating view conflicts caused by inconsistent style information. Additionally, based on the geometric information, we introduce a geometry-aware mask to prevent redundant optimization in multi-view overlapping regions, further improving multi-view consistency. In summary, our key contributions are as follows: 
\begin{itemize}
    \item To the best of our knowledge, DiffStyle3D is the first paradigm to perform 3DGS stylization by optimizing directly in the latent space of a diffusion model.
    \item We propose an Attention-Aware Loss that enables style transfer while preserving the original content.
    \item We propose Geometry-Guided Multi-View Consistency to mitigate multi-view inconsistency.
    \item Extensive experiments demonstrate that our method outperforms existing state-of-the-art approaches in both qualitative and quantitative evaluations.
\end{itemize}
\section{Related Work}
\label{sec:relate_work}
\noindent \textbf{2D Style Transfer}. Style transfer has remained a central topic in generative visual research, aiming to map the stylistic characteristics of a reference image onto a content image. The early pioneering approach~\cite{gatys2016image} achieved neural style transfer by minimizing the distance between Gram matrices derived from VGG features, which motivated extensive follow-up works~\cite{heitz2021sliced,risser2017stable,vacher2020texture}. With the rapid development of diffusion models, existing methods have increasingly relied on this framework, leading to the emergence of numerous fine-tuning-based approaches~\cite{zhou2025attention,ye2023ip,xing2024csgo,yang2025splitflux} and training-free methods~\cite{he2024freestyle,xu2024freetuner,wang2024instantstyle}. Compared to traditional VGG-feature-based methods, diffusion-driven approaches achieve higher content fidelity and improved stylistic expressiveness. Therefore, we propose DiffStyle3D, a fully diffusion-based framework.

\begin{figure*}[h!]
    \centering
    \includegraphics[width=\linewidth]{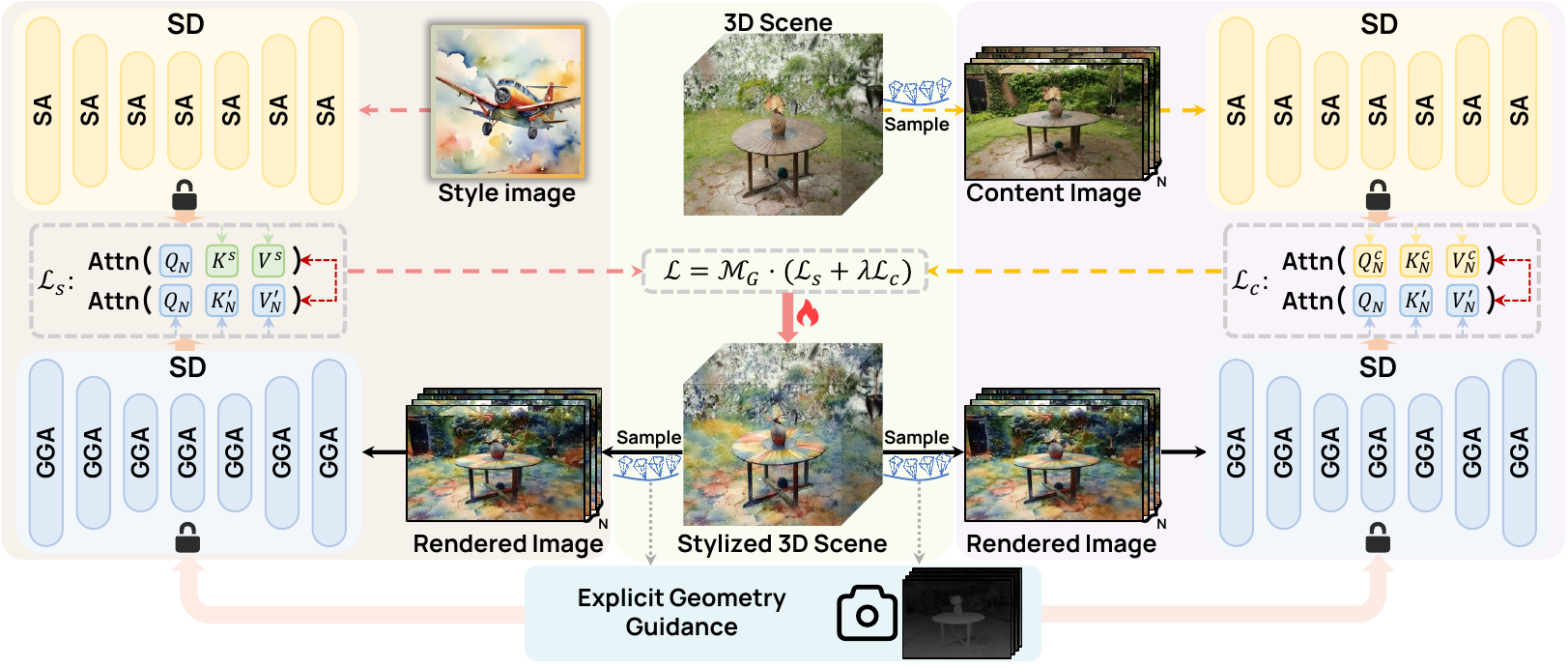}
    \caption{Overview of DiffStyle3D. We introduce an Attention-Aware Loss that enables style transfer while preserving content. To model multi-view correspondences, we derive a explicit geometry guidance from camera parameters and depth maps and incorporate it into Self-Attention (SA) to form Geometry-Guided Attention (GGA). Additionally, a geometry-aware mask $\mathcal{M}_G$ restricts optimization to non-overlapping regions, further improving multi-view consistency.}
    \label{fig:model}
    \vspace{5pt}
\end{figure*}
\noindent \textbf{Attention Control}. As one of the most advanced paradigms in generative modeling, diffusion models~\cite{ho2020denoising,nichol2021improved,rombach2022high} have demonstrated remarkable capabilities across both 2D and 3D domains~\cite{podell2023sdxl,liu2023zero,shi2023mvdream}. At the heart of these models lies the attention mechanism, which has been extensively explored in recent research~\cite{alaluf2024cross,cao2023masactrl}. By imposing various forms of control on self-attention and cross-attention modules, existing methods have achieved superior performance in tasks such as content editing~\cite{yang2025prompt,hertz2022prompt} and style transfer~\cite{hertz2024style,chung2024style}. Based on the self-attention mechanism, we propose an Attention-Aware Loss to effectively transfer stylistic information while preserving the original content.

\noindent \textbf{3DGS Style Transfer}. Following a trajectory similar to that of 2D style transfer, style transfer methods for 3D Gaussian Splatting can be broadly categorized into three groups. Most existing approaches are VGG-based methods, which optimize the 3D scene using feature matching losses~\cite{StyleGaussian,saroha2024gaussian}, multi-scale losses~\cite{galerne2025sgsst}, or nearest-neighbor feature matching losses~\cite{Stylesplat,zhang2024stylizedgs}. Another line of CLIP-based methods~\cite{howil2025clipgaussian,kovacs2024g} achieves stylization by aligning representations in the CLIP embedding space. More recently, diffusion-based methods~\cite{zhuang2025styleme3d} have been explored for 3D stylization, either by generating stylized 2D image supervision using diffusion models~\cite{ArtNVG,InstantStyleGaussian}, which essentially reformulates the problem as a 2D style transfer task, or by directly distilling diffusion models into the 3D representation~\cite{yang2025fantasystyle}. However, these approaches struggle to effectively establish multi-view consistency. In contrast, we introduce Geometry-Guided Multi-View Consistency to enforce cross-view consistency.

\section{Preliminary}
\noindent \textbf{Self-Attention.} Given an input sequence $X \in \mathbb{R}^{n \times d}$, the self-attention mechanism projects each token into a triplet of latent representations consisting of query $Q$, key $K$, and value $V$ through learned linear transformations. The attention output is computed as
\begin{equation}
\mathrm{Attn}(Q,K,V)=\mathrm{Softmax}\!\left(\frac{QK^\top}{\sqrt{d_k}}\right)V,
\label{eq:attn}
\end{equation}
where $d_k$ is the scaling factor. This mechanism allows each token to aggregate information from others according to pairwise relevance, capturing contextual and content-dependent interactions.

\noindent \textbf{3D Gaussian Splatting.}  
3DGS represents a 3D scene as a set of $M$ anisotropic Gaussians, formulated as:
\begin{equation}
\min_\Theta\frac{1}{N}\sum_{i=1}^N \mathcal{L}(\mathcal{R}(C_i;\Theta),I_i^{gt}),
\label{eq:gs}
\end{equation}
where $\Theta = \{(\mu_{m}, \Sigma_{m}, \alpha_{m}, \mathcal{C}_{m})\}^{M}_{m=1}$ denotes the 3D Gaussian parameters. Here, $\mu_{m}$, $\Sigma_{m}$, $\alpha_{m}$, and $\mathcal{C}_{m}$ represent the mean position, covariance matrix, opacity, and spherical harmonics (SH) coefficients for color, respectively. $C_i$ denotes the $i$-th camera parameters, $\mathcal{R}(\cdot)$ the rasterization-based renderer, and $I_i^{gt}$ the corresponding ground-truth image.

\section{Method}
Given a style image $I^{s}$, we aim to transfer its style to the 3D scene while preserving the original content. Therefore, we only optimize the color-related parameters. We propose DiffStyle3D (Fig.~\ref{fig:model}), which employs an Attention-Aware Loss for 3D style transfer via direct latent-space optimization (Sec.~\ref{sec:al}). To capture cross-view correspondences and improve multi-view consistency, we introduce Geometry-Guided Multi-View Consistency (Sec.~\ref{sec:gga}).

\subsection{Attention-Aware Loss}
\label{sec:al}
Previous diffusion model–based approaches for 3D style transfer~\cite{yang2025fantasystyle, zhuang2025styleme3d} typically optimize 3D representations by following the predicted denoising directions. However, this optimization strategy often leads to instability and overly smooth results, limiting their effectiveness in high-quality 3D stylization. Drawing inspiration from inference-time optimization techniques~\cite{chen2024training, ding2024training, shi2024dragdiffusion}, we introduce an Attention-Aware Loss that establishes a new 3D stylization paradigm through direct optimization in the latent space of diffusion models. By aligning representations in the self-attention feature space, our method enables accurate style transfer while preserving original content, without relying on unstable denoising guidance.
\begin{figure}[]
    \centering
    \includegraphics[width=\linewidth]{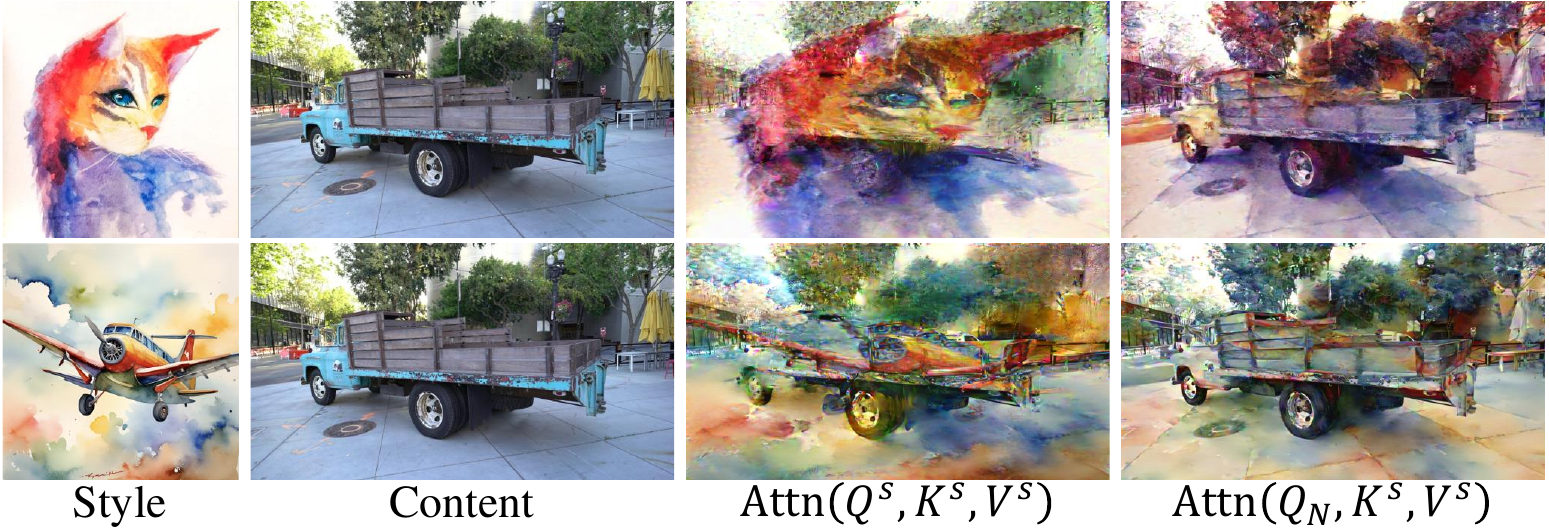}
    \caption{Results with different stylization signals. We conduct experiments using a fixed viewpoint of the 3D scene. Directly using the attention outputs of the style image as stylization signals leads to severe content leakage.}
    \label{fig:attn}
    % \vspace{-15pt}
\end{figure}

Specifically, given a 3D Gaussian scene and a style image $I^s$, we sample $N$ cameras $C_{N} = \{c_1, c_2, \ldots, c_N\}$ in each batch to render the scene, producing a set of rendered images $I_N$ along with their corresponding original content images $I_N^c$. These images are then fed into a diffusion model to extract features, which can be formulated as follows:
\begin{equation}
\begin{aligned}
z_N &= E(I_N),     & z_N^c   &= E(I_N^c), & z^s   &= E(I^s), \\
h_N &\coloneqq \epsilon_\theta(z_N), 
& h_N^{c} &\coloneqq \epsilon_\theta(z_N^c), 
& h^{s}&\coloneqq \epsilon_\theta(z^s),
\end{aligned}
\end{equation}
where $E(\cdot)$ denotes the VAE encoder, $\epsilon_\theta$ represents the UNet. $h_N$, $h_N^{c}$, and $h^{s}$ denote the features extracted from specific layers of the UNet, which are used to compute the style and content losses. 

\noindent \textbf{Style Loss.} Controlling self-attention has been widely adopted in style transfer tasks, motivating us to design our loss function around the self-attention mechanism. A straightforward solution is to directly align the attention outputs of the style image and the rendered image to achieve style transfer. However, such a strategy often leads to severe content leakage from the style image, as illustrated in Fig.~\ref{fig:attn}. To address this issue, we inject stylistic semantics by combining the key (K) and value (V) from the style image with the query (Q) from the rendered image and use the resulting attention output as the stylization signal. Formally, we first extract the $Q_N$, $K_N$, and $V_N$ from the self-attention layers of the rendered image to compute the attention output, which are centered to zero mean and then normalized:
\begin{equation}
\begin{aligned}
% \text{Attn}_{N}= \frac{\text{Attn}(Q_{N},K_{N},V_{N})-\mu(\text{Attn}(Q_{N},K_{N},V_{N}))}{\|\text{Attn}(Q_{N},K_{N},V_{N})-\mu(\text{Attn}(Q_{N},K_{N},V_{N}))\|_2},
\widehat{\mathcal{A}}_N = \frac{\mathcal{A}_N - \mu(\mathcal{A}_N)}
{\left\|\mathcal{A}_N - \mu(\mathcal{A}_N)\right\|_2},\mathcal{A}_N = \text{Attn}(Q_N, K_N, V_N),
\label{eq:op}
\end{aligned}
\end{equation}
where $\mu(\cdot)$ denotes the mean over channels. Meanwhile, $K^s$ and $V^s$ from the style image are integrated with $Q_N$ to inject style semantics:
\begin{equation}
\begin{aligned}
\widehat{\mathcal{A}}^s = \frac{\mathcal{A}^s - \mu(\mathcal{A}^s)}
{\left\|\mathcal{A}^s- \mu(\mathcal{A}^s)\right\|_2}, \mathcal{A}^s = \text{Attn}(Q_N, K^s, V^s).
\label{eq:style}
\end{aligned}
\end{equation}
Finally, style guidance is achieved by minimizing the distance between the two representations:
\begin{equation}
\begin{aligned}
\mathcal{L}_{s}&=\left\|\widehat{\mathcal{A}}_N - \widehat{\mathcal{A}}^s \right\|_2^2.
\label{eq:l_s}
\end{aligned}
\end{equation}
\begin{figure}[]
    \centering
    \includegraphics[width=\linewidth]{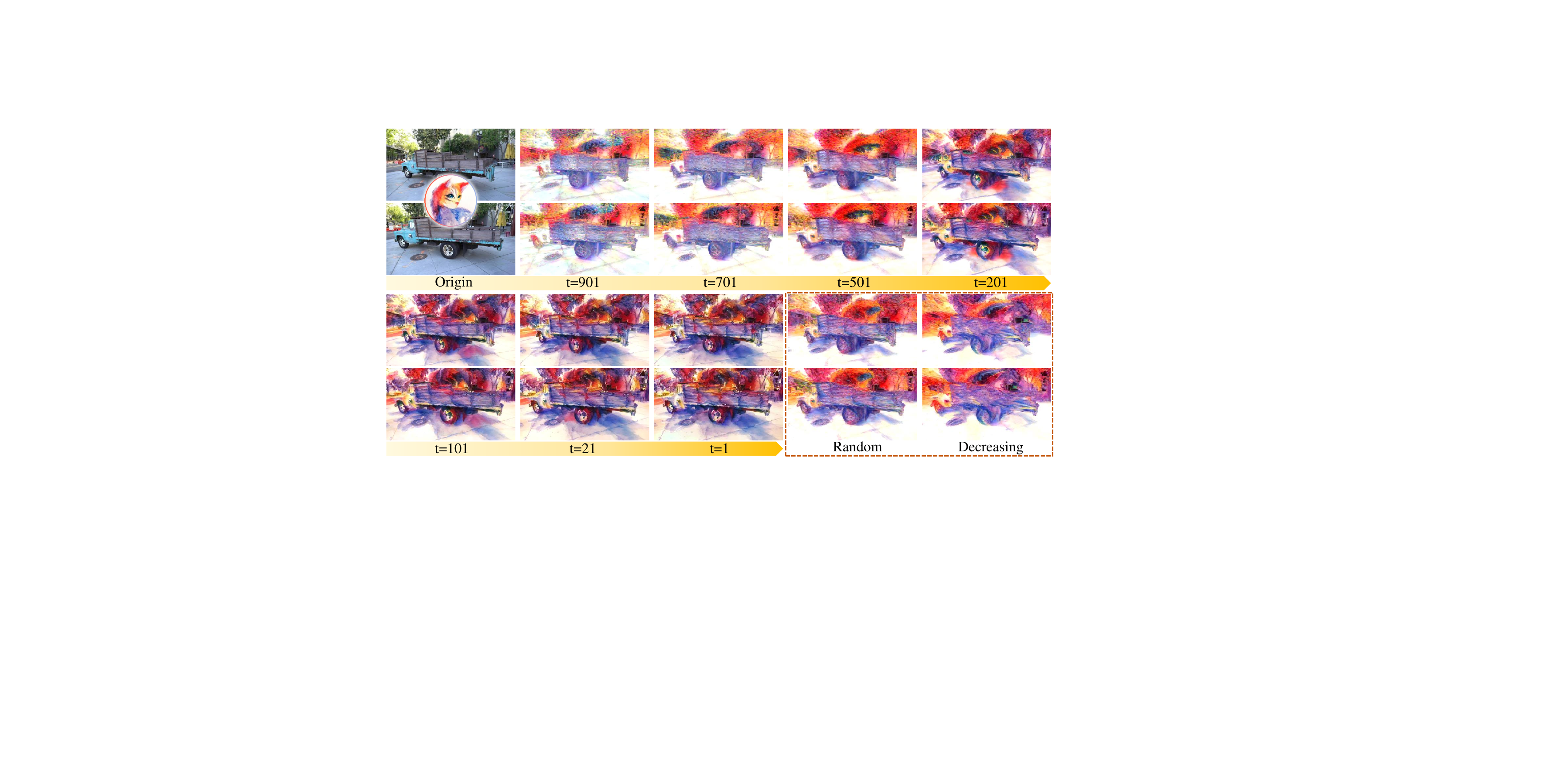}
    \caption{Results obtained using different timestep during optimization. \textit{Random} denotes randomly sampled timestep throughout the optimization process, while \textit{decreasing} simulates the diffusion process by progressively decreasing the time step from $T$ to 0.}
    \label{fig:timestep}
    % \vspace{-15pt}
\end{figure}
Directly applying $\ell_1$ or $\ell_2$ loss on $\mathcal{A}^s$ and $\mathcal{A}_N$ focuses on the absolute values of the features, which can slow down training. By centering and normalizing features before computing the loss, we emphasize the consistency of their direction and patterns, enabling a more effective style transfer.

\noindent \textbf{Content Loss.} A central challenge in style transfer is to effectively apply the target style while preserving the original content. To prevent over-stylization that could distort local features or semantic structures in the rendered image, we design a content loss. Similar to the style loss, the content loss is defined based on self-attention. It preserves the original content by minimizing the distance between the attention representations of the content image and the rendered image. Formally, it can be expressed as follows:
\begin{equation}
\begin{aligned}
\widehat{\mathcal{A}}^c_N = \frac{\mathcal{A}^c_N - \mu(\mathcal{A}^c_N)}
{\left\|\mathcal{A}^c_N - \mu(\mathcal{A}^c_N)\right\|_2},\mathcal{A}^c_N = \text{Attn}(Q^c_N, K^c_N, V^c_N),
\label{eq:content}
\end{aligned}
\end{equation}
\begin{equation}
\begin{aligned}
% \nonumber
\mathcal{L}_{c}=\left\|\widehat{\mathcal{A}}_N - \widehat{\mathcal{A}}^c_N \right\|_2^2.
\label{eq:l_c}
\end{aligned}
\end{equation}

\noindent \textbf{Timestep Choice.} As illustrated in Fig.~\ref{fig:timestep}, we analyze the effect of different fixed diffusion timesteps during optimization. Larger timesteps introduce increased noise and result in blurred stylization, while smaller timesteps better preserve fine-grained stylistic details and brushstroke textures. We further consider \textit{random} and \textit{decreasing} timestep strategies commonly used in 3D generation. However, since these strategies still involve large timesteps, they tend to introduce blurring artifacts. As a result, we adopt a fixed timestep of $t = 1$ as a key choice in our method.

\subsection{Geometry-Guided Multi-View Consistency}
\label{sec:gga}
Although the Attention-Aware Loss achieves promising results in style transfer, the same object may still receive inconsistent stylistic representations across different views, leading to noticeable cross-view inconsistency. To address this issue, we propose Geometry-Guided Multi-View Consistency. Unlike VGG-based~\cite{galerne2025sgsst} and CLIP-based methods~\cite{howil2025clipgaussian}, which struggle to model such cross-view constraints within the model, our approach leverages the intrinsic self-attention mechanism of diffusion models to capture correlations across different views, thereby improving multi-view consistency.

\noindent \textbf{Explicit Geometry Guidance}. In our framework, we optimize only color-related parameters, ensuring geometric invariance of the 3D Gaussians and a fixed depth map $D_b$ for any given viewpoint $b$. This stability allows us to explicitly establish geometric correspondences across views using known camera intrinsics and poses. For a pixel $\mathbf{p}$ in the reference view $b$, its corresponding sampling coordinate in source view $j$, denoted as $\mathbf{g}_{b \leftarrow j}(\mathbf{p})$, is derived via back-projection and re-projection:
\begin{equation}
\mathbf{g}_{b \leftarrow j}(\mathbf{p})
=
\Pi\!\left(
\mathbf{K}_j \,
\mathbf{T}_{j}^{w2c}
\mathbf{T}_{b}^{c2w}
D_b(\mathbf{p})
\mathbf{K}_b^{-1}
\tilde{\mathbf{p}}
\right),
\label{eq:grid}
\end{equation}
where $\tilde{\mathbf{p}}$ represents the homogeneous coordinates of $\mathbf{p}$, while $\mathbf{K}$ and $\mathbf{T}$ denote camera intrinsics and extrinsics. $w2c$ and $c2w$ denote the world-to-camera and camera-to-world transformations, respectively. $\Pi(\cdot)$ represents perspective projection followed by normalization to the $[-1,1]$ sampling space. To account for occlusions and boundaries, we define a visibility mask:
\begin{equation}
\mathbf{v}_{b \leftarrow j}(\mathbf{p})
=
\mathbf{1}
\big(
\text{inFront}_j(\mathbf{p})
\;\wedge\;
\mathbf{g}_{b \leftarrow j}(\mathbf{p}) \in \Omega
\big),
\label{eq:visible}
\end{equation}
where $\mathbf{1}(\cdot)$ denotes the indicator function, $\Omega$ denotes the valid image domain, and $\text{inFront}_j(\cdot)$ enforces that the re-projected point has a positive depth in view $j$.

\noindent \textbf{Geometry-Guided Attention.} We integrate the obtained sampling grids and visibility masks into the self-attention mechanism of the diffusion model to explicitly model correspondences across multiple views, thereby strengthening cross-view consistency constraints. Specifically, we augment $K$ and $V$ of the reference view by warping features from all other views within the batch. For a batch of $N$ views, $K'_b$ and $V'_b$ for view $b$ are formulated as:
\begin{equation}
\begin{aligned}
K'_b
&=
[
K_b;\;
\{\, \mathcal{W}_{b\leftarrow j}(K_j)
\mid j \in \{0,\dots,N-1\},\ j \neq b \,\}
],\\
V'_b
&=
[
V_b;\;
\{\, \mathcal{W}_{b\leftarrow j}(V_j)
\mid j \in \{0,\dots,N-1\},\ j \neq b \,\}
],
\label{eq:new_attn_kv}
\end{aligned}
\end{equation}
where $[\cdot\,;\,\cdot]$ denotes the concatenation and $\mathcal{W}_{b\leftarrow j}(\cdot)$ represents the bilinear warping operator guided by $\mathbf{g}_{b \leftarrow j}$. By rewriting $\mathcal{A}_N$ in Eq.~\ref{eq:op}, the Geometry-Guided Attention (GGA) formula is defined as:
\begin{equation}
\mathrm{Attn}(Q_N, K'_N, V'_N)
= \mathrm{Softmax}\!\left(
\frac{Q_N{K'_N}^\top}{\sqrt{d_k}} + \mathbf{M}_\mathbf{v}
\right)V'_N,
\label{eq:new_attn}
\end{equation}
where $K_N^{\prime}=\{K_b^{\prime}\}_{b=0}^{N-1}$, $V_N^{\prime}=\{V_b^{\prime}\}_{b=0}^{N-1}$. $\mathbf{M}_\mathbf{v}$ consists of visibility masks $\mathbf{v}$ and serves as the attention mask to prevent erroneous feature aggregation from occluded regions.

\noindent \textbf{Geometry-Aware Mask.} To avoid redundant optimization over multi-view overlapping regions, we introduce a geometry-aware mask $\mathcal{M}_G$, which is defined as follows:
\begin{equation}
\mathcal{M}_G
\coloneqq
\left\{
\mathbf{1}
\!\left(
\forall\, j < b,\;
\mathbf{v}_{b \leftarrow j}(\mathbf{p}) = 0
\right)
\right\}_{b,\mathbf{p}}.
\end{equation}
\(\mathcal{M}_G\) is a collection of masks, one for each view. For a given view \(b\), its mask retains only the pixels that have not been observed in any previous view \(j<b\), setting them to 1. Consequently, \(\mathcal{M}_G\) represents the non-overlapping regions across all views in the current batch, which further improves multi-view consistency.

\noindent \textbf{Optimization Objective.} We extract features from all self-attention layers of the diffusion model. The final loss is defined as follows:
\begin{equation}
\begin{aligned}
\mathcal{L}=\mathcal{M}_G \cdot (\mathcal{L}_{s} +\lambda\mathcal{L}_{c}).
\end{aligned}
\end{equation}
where $\lambda$ is a scaling factor that controls the strength of content preservation during style transfer. Our method is detailed in Alg.~\ref{alg:1}.
\begin{algorithm}[h]
\caption{DiffStyle3D}
\label{alg:1}
\begin{algorithmic}[1]
\REQUIRE VAE encoder $E$, diffusion model $\epsilon_\theta$, GGA-integrated diffusion model $\epsilon'_\theta$, training iterations $S$, fixed timestep $t$, $N$ views per batch, style image $I^s$.
\FOR{$s = 1$ to $S$}
    \STATE \textbf{Sample:} $I_{N} = \mathcal{R}(C_N;\Theta)$, $I_N^c$, depth maps $D_N$, camera $C_N$ intrinsics and extrinsics.
    \STATE \textbf{Explicit Geometry Guidance:} $\mathbf{g}$ and $\mathbf{v}$ defined by Eq.~\ref{eq:grid}, Eq.~\ref{eq:visible}, geometry-aware mask $\mathbf{M}_G$.
    \STATE $z_N,\; z_N^c,\; z^s \; \leftarrow \; E(I_N),\; E(I_N^c),\; E(I^s)$
    \STATE
    $\begin{alignedat}{3}
    &\widehat{\mathcal{A}}^c_N, Q_N \;\leftarrow\; h_N^{c} \;\coloneqq\; \epsilon_\theta(z_N^c,t), & \text{in Eq.~\ref{eq:content}}& \\
    &\widehat{\mathcal{A}}^s   \;\leftarrow\; \{Q_N,\quad h^{s}     \;\coloneqq\;\epsilon_\theta(z^s,t)\},   &\text{in Eq.~\ref{eq:style}}& \\
    &\widehat{\mathcal{A}}_N   \;\leftarrow\; h_N       \;\coloneqq\; \epsilon'_\theta(z_N,t),  & \text{in Eq.}~\ref{eq:op},~\ref{eq:new_attn_kv},~\ref{eq:new_attn}&
    \end{alignedat}$
    \STATE $\mathcal{L} = \mathcal{M}_G(\mathcal{L}_{s}(\widehat{\mathcal{A}}_N,\widehat{\mathcal{A}}^s) + \lambda\mathcal{L}_{c}(\widehat{\mathcal{A}}_N,\widehat{\mathcal{A}}^c_N))$ Eq.~\ref{eq:l_s},~\ref{eq:l_c}
    \STATE Compute $\nabla_{z_N}\mathcal{L}$
    \STATE Update 3D Gaussians using $\nabla_{z_N}\mathcal{L}$
\ENDFOR
\ENSURE 3D Stylized Scene.
\end{algorithmic}
\end{algorithm}

\begin{table*}[!t]
\centering
\caption{Quantitative comparison of different methods in 3DGS style transfer. \colorbox{orange!15}{\textbf{Bold}}: best; \colorbox{orange!7}{\underline{underline}}: second best.}
\begin{adjustbox}{max width=\textwidth} 
\begin{tabular}{cccccccccccc}
\hline
\multirow{2}{*}{Method} & \multirow{2}{*}{CLIP-S $\uparrow$}& \multirow{2}{*}{CLIP-C $\uparrow$}& \multirow{2}{*}{CLIP-CONS $\uparrow$}& \multirow{2}{*}{CLIP-F} & \multirow{2}{*}{$S_{vgg} \downarrow$}  & \multirow{2}{*}{$\text{FID}$ $\downarrow$}            & \multicolumn{2}{c}{Short-range consistency} & \multicolumn{2}{c}{Long-range consistency} & \multicolumn{1}{c}{Per-Instance}\\ \cline{8-11} 
            & & &            &              &           &                                   & LPIPS  $\downarrow$              & RMSE  $\downarrow$               & LPIPS   $\downarrow$             & RMSE  $\downarrow$     & training time$\downarrow$         \\ \hline
% ARF                     &    &   &               &              &              &              &            \\
\multicolumn{12}{c}{\cellcolor{blue!10}Scene-level} \\ \hline
StyleGaussian       & 0.64& 0.61 & 0.032   &  1.02   &  25.81 &     334.9            &   0.088        &     0.110        &      \cellcolor{orange!7}\underline{0.151}        &    0.160 & $\sim$21min        \\
SGSST           &0.65&0.64 &0.066        & 1.01  &  \cellcolor{orange!15}\textbf{20.60}  &      289.1        &    0.086         &    0.107       &  0.159          &     0.182     & $\sim$40min \\ 
FantasyStyle       &0.64&\cellcolor{orange!7}\underline{0.67}& \cellcolor{orange!7}\underline{0.084}             &\cellcolor{orange!7}\underline{1.01}  & 26.88 & \cellcolor{orange!7}\underline{228.6} &    0.084          & 0.101             &  0.167            &   0.173    & $\sim$34min     \\ 
CLIPGaussian        &\cellcolor{orange!15}\textbf{0.79} & 0.63 &   0.058        & 1.03  & 24.68   &  280.7            &  \cellcolor{orange!7}\underline{0.081}           &      \cellcolor{orange!15}\textbf{0.074}     &   0.161         &   \cellcolor{orange!15}\textbf{0.127}   & \cellcolor{orange!7}\underline{$\sim$16min}     \\ \hline
Ours      &\cellcolor{orange!7}\underline{0.65}&\cellcolor{orange!15}\textbf{0.71}& \cellcolor{orange!15}\textbf{0.128}             &  \cellcolor{orange!15}\textbf{1.00} & \cellcolor{orange!7}\underline{20.63} & \cellcolor{orange!15}\textbf{204.8}               &   \cellcolor{orange!15}\textbf{0.075}           & \cellcolor{orange!7}\underline{0.075}             &     \cellcolor{orange!15}\textbf{0.141}         &   \cellcolor{orange!7}\underline{0.130}  & \cellcolor{orange!15}\textbf{$\sim$16min}       \\ \hline
\multicolumn{12}{c}{\cellcolor{blue!10}Object-level} \\ \hline
SGSST           &0.63&0.75 &0.224        & 1.02  &  \cellcolor{orange!15}\textbf{13.94}  &      333.7        &    0.184         &    0.176       &  0.211          &     0.202     & $\sim$6min \\ 
FantasyStyle       &0.61&\cellcolor{orange!7}\underline{0.79}& 0.358             &\cellcolor{orange!7}\underline{1.02}  & 18.20 & 265.2 &    0.186          & 0.150             &  0.218            &   0.188    & \cellcolor{orange!7}\underline{$\sim$5min}     \\ 
CLIPGaussian        &\cellcolor{orange!15}\textbf{0.72} & 0.79 &   \cellcolor{orange!15}\textbf{0.504}        & 1.03  & 16.70   &  \cellcolor{orange!7}\underline{259.6}           &  \cellcolor{orange!7}\underline{0.177}           &      \cellcolor{orange!15}\textbf{0.111}     &   \cellcolor{orange!7}\underline{0.209}         &   \cellcolor{orange!15}\textbf{0.155}   & $\sim$13min     \\ \hline
Ours      &\cellcolor{orange!7}\underline{0.63}&\cellcolor{orange!15}\textbf{0.84}& \cellcolor{orange!7}\underline{0.412}             &  \cellcolor{orange!15}\textbf{1.01} & \cellcolor{orange!7}\underline{16.44} & \cellcolor{orange!15}\textbf{255.8}               &   \cellcolor{orange!15}\textbf{0.176}           & \cellcolor{orange!7}\underline{0.119}             &     \cellcolor{orange!15}\textbf{0.208}         &   \cellcolor{orange!7}\underline{0.161}  & \cellcolor{orange!15}\textbf{$\sim$4min}       \\ \hline
\end{tabular}
\end{adjustbox}
\label{tab:comparison}
\end{table*}

\begin{figure*}[]
    \vspace{-8pt}
    \centering
    \includegraphics[width=\linewidth]{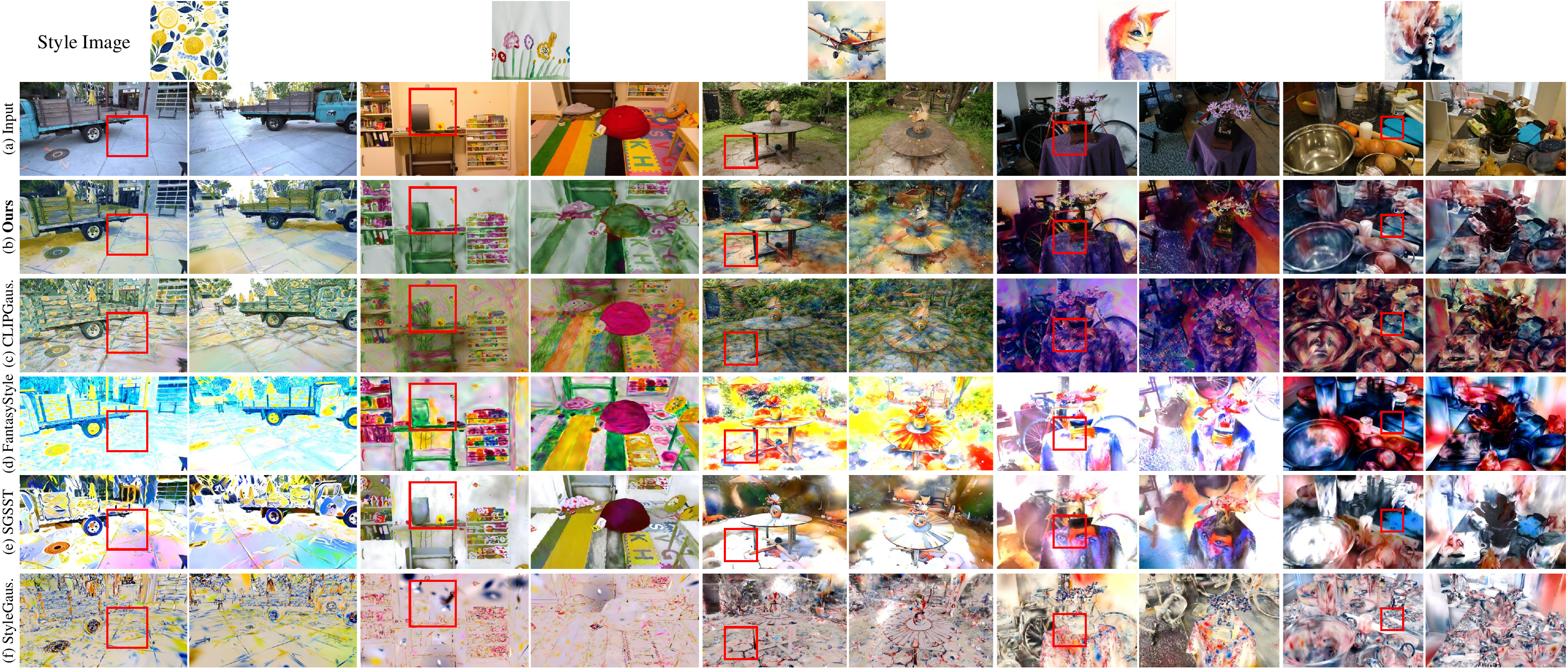}
    \caption{Qualitative comparison of different methods on scene-level datasets. Our approach achieves superior style transfer while better preserving the original content. The red boxes highlight clear differences, the details of which are further compared in Fig.~\ref{fig:comparison_detail}.}
    \label{fig:comparison}
    \vspace{-15pt}
\end{figure*}

\section{Experiments}
\noindent \textbf{Datasets.} We select 8 scenes from the Tandt DB dataset~\cite{kerbl20233d} and the Mip-NeRF 360 dataset~\cite{barron2022mip}. Each scene is stylized using 14 different style images, resulting in a total of 112 stylization experiments. In addition, we employ SAM3D~\cite{chen2025sam} to extract 10 individual objects, producing 140 object-level stylization results in total. We comprehensively evaluate our method at both the scene-level and the object-level across diverse artistic domains.
\begin{figure*}[]
    \centering
    \includegraphics[width=\linewidth]{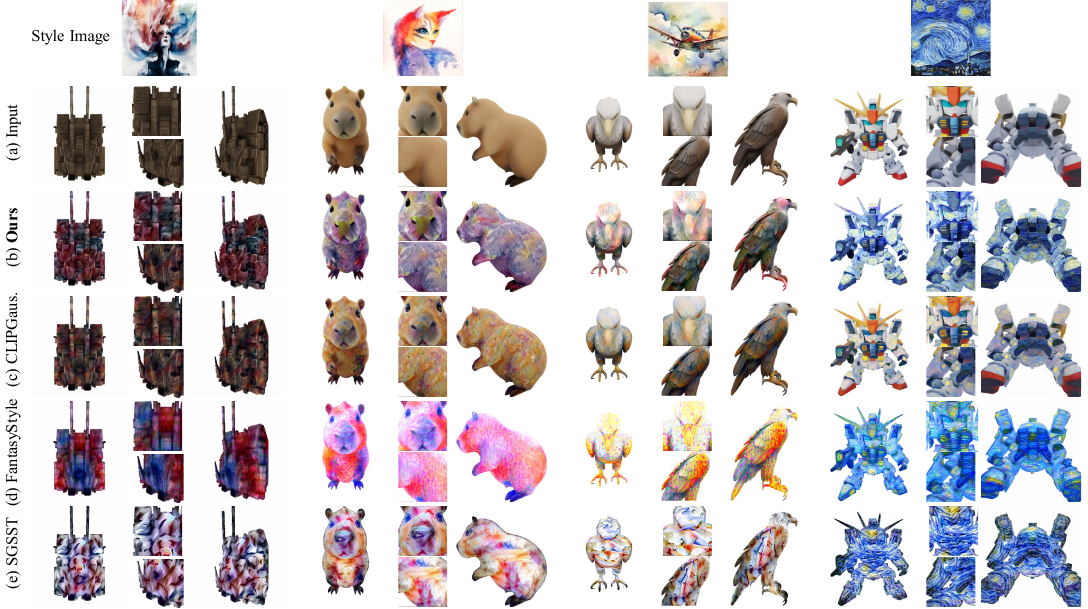}
    \caption{Qualitative comparison of different methods on object-level datasets. Other methods often suffer from over-stylization and content leakage from the style image. In contrast, our approach avoids these issues, achieving superior visual quality in style transfer.}
    \label{fig:objs}
    \vspace{-15pt}
\end{figure*}
\begin{figure}[t]
    \centering
    \includegraphics[width=\linewidth]{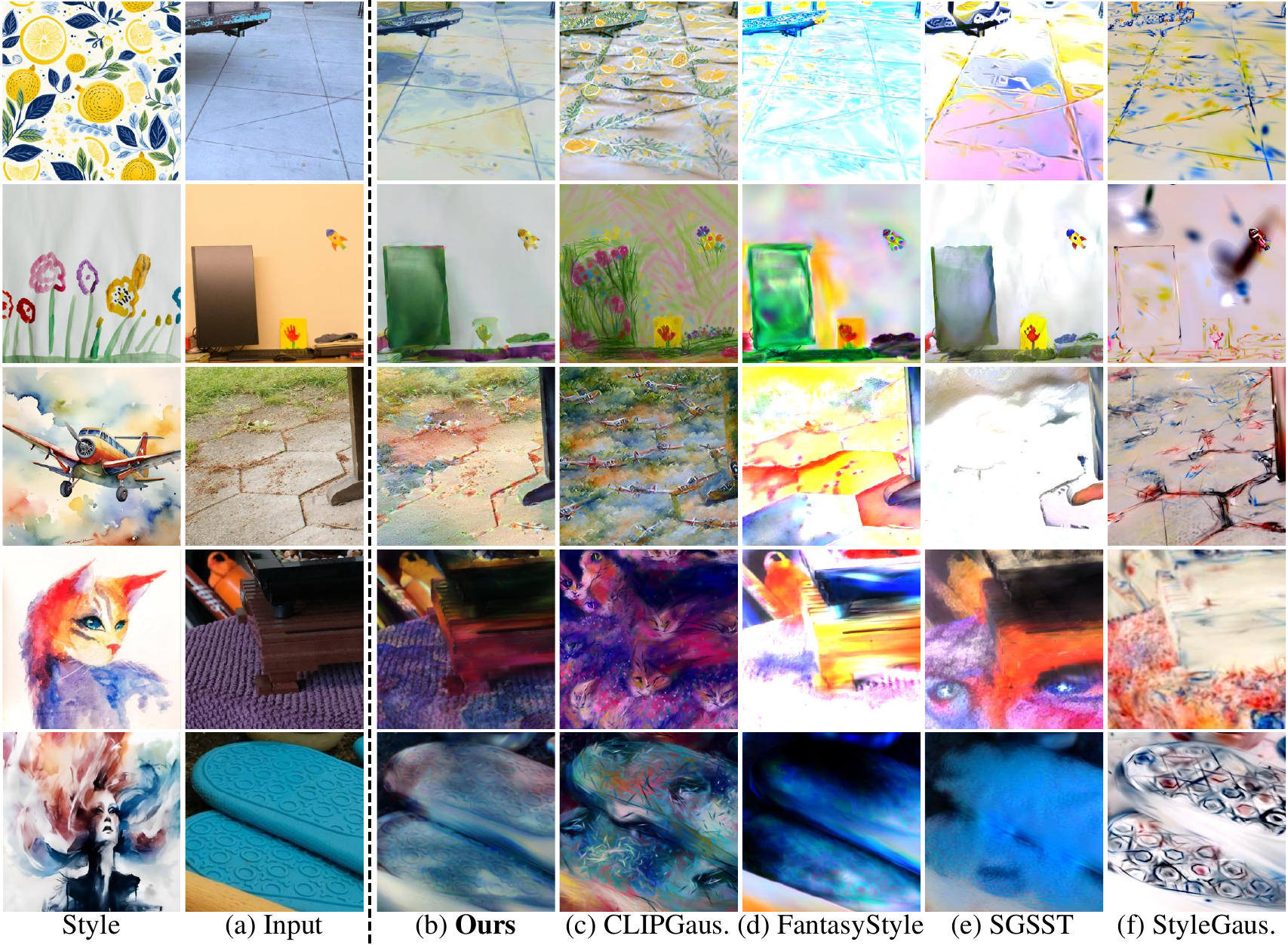}
    \caption{Detailed qualitative comparison of different methods. Our approach adheres more closely to the target style and avoids content leakage from the style image, compared to existing methods. Zoom in for better view.}
    \label{fig:comparison_detail}
    % \vspace{-20pt}
\end{figure}

% LPIPS~\cite{zhang2018unreasonable}
\noindent \textbf{Metrics.} We evaluate content preservation using CLIP-C~\cite{radford2021learning} and FID, while CLIP-S is used to assess style transfer quality. CLIP-CONS and CLIP-F~\cite{howil2025clipgaussian} are employed to measure semantic temporal consistency, where CLIP-F values closer to 1 indicate better consistency. To evaluate overall transfer quality, we define the $S_{vgg}$, which is computed using features extracted from VGG19~\cite{simonyan2014very}. Furthermore, LPIPS and RMSE are used to measure short-term and long-term multi-view consistency~\cite{StyleGaussian}, respectively.

\noindent \textbf{Comparison Methods.} We compare our method with recent state-of-the-art approaches, including VGG-based methods (StyleGaussian~\cite{StyleGaussian}, SGSST~\cite{galerne2025sgsst}), CLIP-based methods (CLIPGaussian~\cite{howil2025clipgaussian}), and diffusion-based methods (FantasyStyle~\cite{yang2025fantasystyle}).

\noindent \textbf{Implementation Details}. We adopt Stable Diffusion 1.5~\cite{rombach2022high} as our base model. We fix the timestep to $t = 1$ and extract self-attention features from all blocks for loss computation. For each batch, we use $N = 4$ views and set $\lambda = 0.1$. All experiments are conducted on a single NVIDIA L20 (48G) GPU.

\subsection{Comparison Results}
\noindent \textbf{Quantitative Comparisons}. As shown in Tab.~\ref{tab:comparison}, we comprehensively evaluate our method across multiple metrics covering style transfer quality, content preservation, and multi-view consistency. Overall, our method achieves the best performance across these aspects. Specifically, CLIPGaussian attains very high scores on CLIP-S, as it directly optimizes style transfer using CLIP-extracted features; however, this comes at the cost of inferior content preservation. Similarly, SGSST performs well on the $S_{vgg}$ metric due to explicitly optimizing this objective. Despite not being tailored to any single metric, our method achieves competitive or superior performance across all style and content metrics. Notably, our method demonstrates significant improvements on multi-view consistency metrics, such as CLIP-CONS and LPIPS, highlighting its effectiveness in enforcing cross-view coherence. In addition, we compare the training time of different methods. Although diffusion models are substantially larger than CLIP and VGG, our method achieves training time comparable to CLIPGaussian and substantially outperforms other methods.
\begin{figure}[t]
    \centering
    \includegraphics[width=\linewidth]{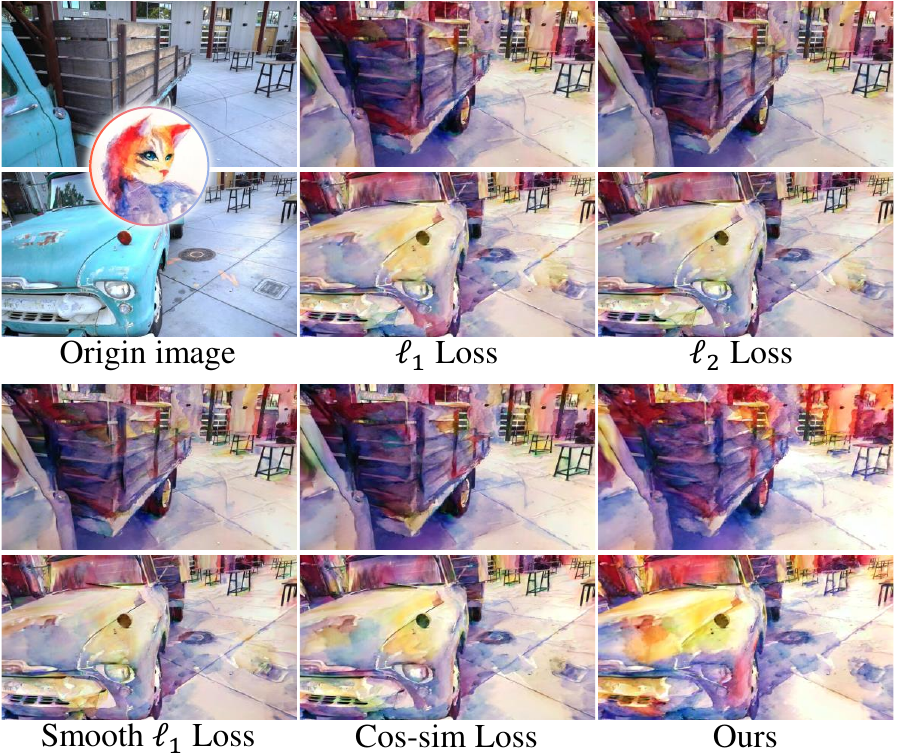}
    \caption{Results of directly applying the losses defined in Eq.~\ref{eq:op},~\ref{eq:style} and~\ref{eq:content} without centering and normalization. In contrast, our method achieves faster style transfer.}
    \label{fig:loss}
    % \vspace{-22pt}
\end{figure}

\begin{table*}[t]
\centering
\caption{Quantitative results of the ablation study on the effect of Geometry-Guided Multi-View Consistency.}
\label{tab:ablation_gga}
\begin{tabular}{lcccccc}
\hline
Method 
& CLIP-CONS$\uparrow$
& CLIP-F 
& \multicolumn{2}{c}{Short-range Consistency} 
& \multicolumn{2}{c}{Long-range Consistency} \\ \cline{4-7}
& & & LPIPS $\downarrow$ & RMSE $\downarrow$ 
& LPIPS $\downarrow$ & RMSE $\downarrow$ \\
\hline
w/o GGA & 0.121 & 1.01 & 0.078 & 0.079 & 0.151 & 0.140 \\
w/o $\mathcal{M}_G$ & 0.124 & 1.01 & 0.076 & 0.076 & 0.142 & 0.132 \\
Ours & \textbf{0.128} & \textbf{1.00} & \textbf{0.075} & \textbf{0.075} & \textbf{0.141} & \textbf{0.130} \\
\hline
\end{tabular}
% \vspace{-15pt}
\end{table*}
\noindent \textbf{Qualitative Comparisons}. Fig.~\ref{fig:comparison}, \ref{fig:objs}, and \ref{fig:comparison_detail} present visual comparisons of different methods on both scene-level and object-level datasets. It can be clearly observed that StyleGaussian struggles to faithfully transfer the target style and severely damages the original scene content. SGSST, based on VGG features, suffers from VGG’s limited representational capacity, making it difficult to handle complex style images and resulting in unsatisfactory stylization (e.g., 5th and 6th columns in Fig.~\ref{fig:comparison}). It can also produce noticeable color distortions, such as large pink and blue regions on the ground (1st and 2nd columns), and may cause content leakage from the style image (4th row in Fig.~\ref{fig:comparison_detail}) or over-stylization that obscures the original content (5th row in Fig.~\ref{fig:comparison_detail}). FantasyStyle is able to preserve the original content relatively well; however, as it relies on IP-Adapter for style transfer, it fails to fully align with the target style. As a result, some stylized outputs deviate from the intended style appearance (e.g., the 1st, 2nd, 5th, and 6th columns in Fig.~\ref{fig:comparison}). CLIPGaussian, which performs stylization by aligning CLIP features, suffers from severe content leakage from the style image, as shown in Fig.~\ref{fig:comparison_detail} and Fig.~\ref{fig:objs}. It often introduces explicit semantic elements from the style image, such as cat faces or human eyes, which also explains its superior performance on the CLIP-S metric. In contrast, our approach aligns with the target style more accurately than diffusion-based methods, yielding higher-quality style transfer results. Compared with VGG- and CLIP-based methods, it better preserves the original content and effectively avoids content leakage from the style image. Overall, our method achieves the best visual quality among all compared approaches.

\subsection{Ablation Study}
\noindent \textbf{Attention-Aware Loss.} We conduct an ablation study on the centering and normalization operations in the Attention-Aware Loss under the same optimization settings, with the results shown in Fig.~\ref{fig:loss}. Without centering and normalization, style transfer remains incomplete, as the optimization process overemphasizes absolute feature magnitudes, resulting in slow convergence. In contrast, incorporating centering and normalization shifts the optimization focus toward feature directions, enabling faster convergence and more effective transfer of the target style.

\begin{figure}[t]
    \centering
    \includegraphics[width=\linewidth]{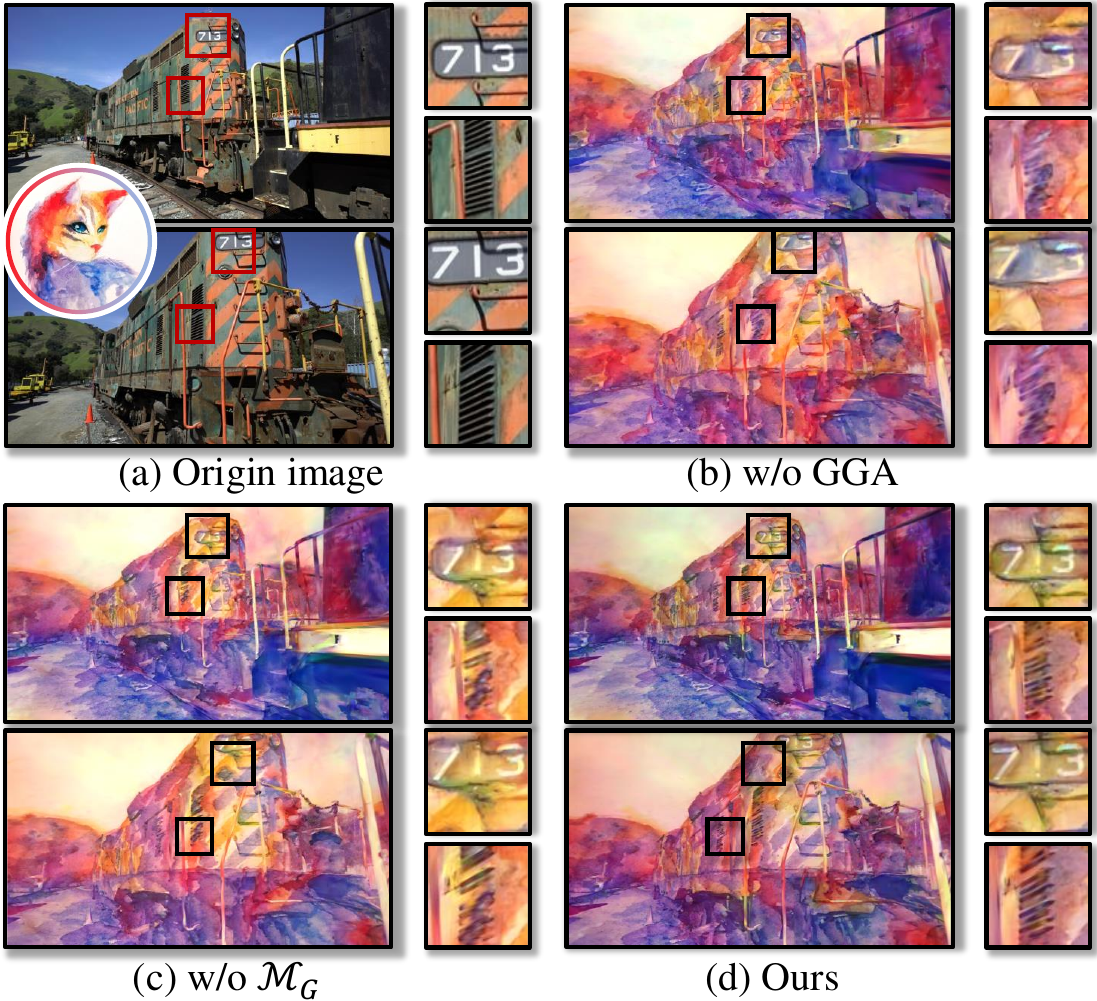}
    \caption{Qualitative results of the ablation study on Geometry-Guided Multi-View Consistency. Zoom in for better view.}
        % \caption{Qualitative results of the ablation study on Geometry-Guided Multi-View ConsistencyGGA and $\mathcal{M}_G$. Zoom in for better view.}
    \label{fig:gga}
    % \vspace{-17pt}
\end{figure}
\noindent \textbf{Geometry-Guided Multi-View Consistency}. We conduct extensive quantitative experiments to evaluate the effectiveness of Geometry-Guided Attention (GGA) in improving multi-view consistency, with the results summarized in Tab.~\ref{tab:ablation_gga}. Our method achieves consistent improvements across all metrics, with particularly notable gains in long-term consistency. We further investigate the role of the geometry-aware mask $\mathcal{M}_G$, which is designed to prevent redundant optimization over geometrically overlapping regions that could otherwise disrupt view consistency. The results show that removing $\mathcal{M}_G$ results in only marginal performance degradation, since GGA already establishes strong multi-view correspondences. This observation further highlights the effectiveness of GGA in enforcing multi-view consistency. The corresponding qualitative results are shown in Fig.~\ref{fig:gga}. Without explicit multi-view modeling, local features of the original content tend to become blurred and overly smoothed. In contrast, we leverage geometric information to model cross-view relationships, preserving sharp local details and coherent structures across different viewpoints, thereby significantly improving visual quality.

% \subsection{Limitation}

\section{Conclusion}
In this work, we propose DiffStyle3D, a novel diffusion-based paradigm for 3DGS stylization that operates directly in the latent space, thereby avoiding unstable denoising guidance. It introduces an Attention-Aware Loss for style transfer and content preservation, and a Geometry-Guided Multi-View Consistency that injects geometric information into self-attention to form Geometry-Guided Attention, enabling cross-view correspondence modeling. Additionally, a geometry-aware mask enhances multi-view consistency by avoiding redundant optimization in overlapping regions. Extensive experiments demonstrate that DiffStyle3D achieves superior stylization quality, better visual quality, and improved content preservation compared to existing methods.

\bibliography{arxiv}

\end{document}